\documentclass[11pt]{article}

\usepackage[preprint]{acl}

\usepackage{times}
\usepackage{latexsym}

\usepackage[T1]{fontenc}

\usepackage[utf8]{inputenc}

\usepackage{microtype}

\usepackage{inconsolata}

\usepackage{graphicx}
\usepackage{booktabs}
\usepackage{xcolor}
\usepackage{array,adjustbox,colortbl}
\usepackage{amsmath,amssymb}

\newcommand{\equalmark}{\textsuperscript{*}}
\newcommand{\leadmark}{\textsuperscript{$\ddagger$}}
\newcommand{\corrmark}{\textsuperscript{$\dagger$}}

\title{MMSearch-OPD: Easy Made Hard, Hard Made Learnable for Long-Horizon Multimodal Search Agents}

\author{
  \begin{minipage}{0.95\textwidth}
  \centering
  \normalsize
  Huanyao Zhang\textsuperscript{1}\equalmark\leadmark\hspace{0.7em}
  Jiepeng Zhou\textsuperscript{2}\equalmark\hspace{0.7em}
  Jiaye Lin\textsuperscript{3}\equalmark\hspace{0.7em}
  Ontai Tang\textsuperscript{4}\\[4pt]
  Jingtian Hao\textsuperscript{1}\quad
  Mugeng Liu\textsuperscript{1}\quad
  Ru Peng\textsuperscript{5}\quad
  Lang Mei\textsuperscript{6}\\[4pt]
  Xiaohan Yu\textsuperscript{6}\quad
  Liyan Liu\textsuperscript{6}\quad
  Chong Chen\textsuperscript{6}\corrmark\quad
  Wentao Zhang\textsuperscript{1}\corrmark\\[8pt]
  \small
  \textsuperscript{1}Peking University\quad
  \textsuperscript{2}HKUST(GZ)\quad
  \textsuperscript{3}Tsinghua University\quad
  \textsuperscript{4}The University of Hong Kong\\[2pt]
  \textsuperscript{5}Zhejiang University\quad
  \textsuperscript{6}Huawei Cloud BU\\[4pt]
  \textsuperscript{*} Equal contribution\quad
  \textsuperscript{$\ddagger$} Project Leader\quad
  \textsuperscript{$\dagger$} Corresponding authors
  \end{minipage}
}

\begin{document}
\maketitle
\begin{abstract}
Multimodal deep search is moving from short queries to long-horizon
investigations that continually acquire evidence. However, training
such agents remains constrained by limited challenging training
questions and insufficient process feedback.
We introduce \textbf{MMSearch-OPD}, combining controllable data
refinement with joint reinforcement learning (RL) and on-policy
distillation (OPD). Our \textbf{Refine} pipeline performs targeted
expansion of reasoning chains derived from existing simple
multimodal questions. Parameterized transformations control
search depth and cross-modal evidence requirements while reusing
existing image--question pairs. For training, answer rewards and
lightweight behavior penalties guide RL, while the reward gap
modulates token-level teacher guidance on imperfect student
trajectories. An exponentially decaying distillation coefficient
reduces teacher dependence as training proceeds.
Experiments show gains from data refinement and further
improvements from joint training after SFT, indicating stronger
long-horizon multimodal deep search capabilities.

\end{abstract}

\section{Introduction}
\label{sec:intro}

\begin{figure*}[t]
  \centering
  \includegraphics[width=0.90\textwidth]{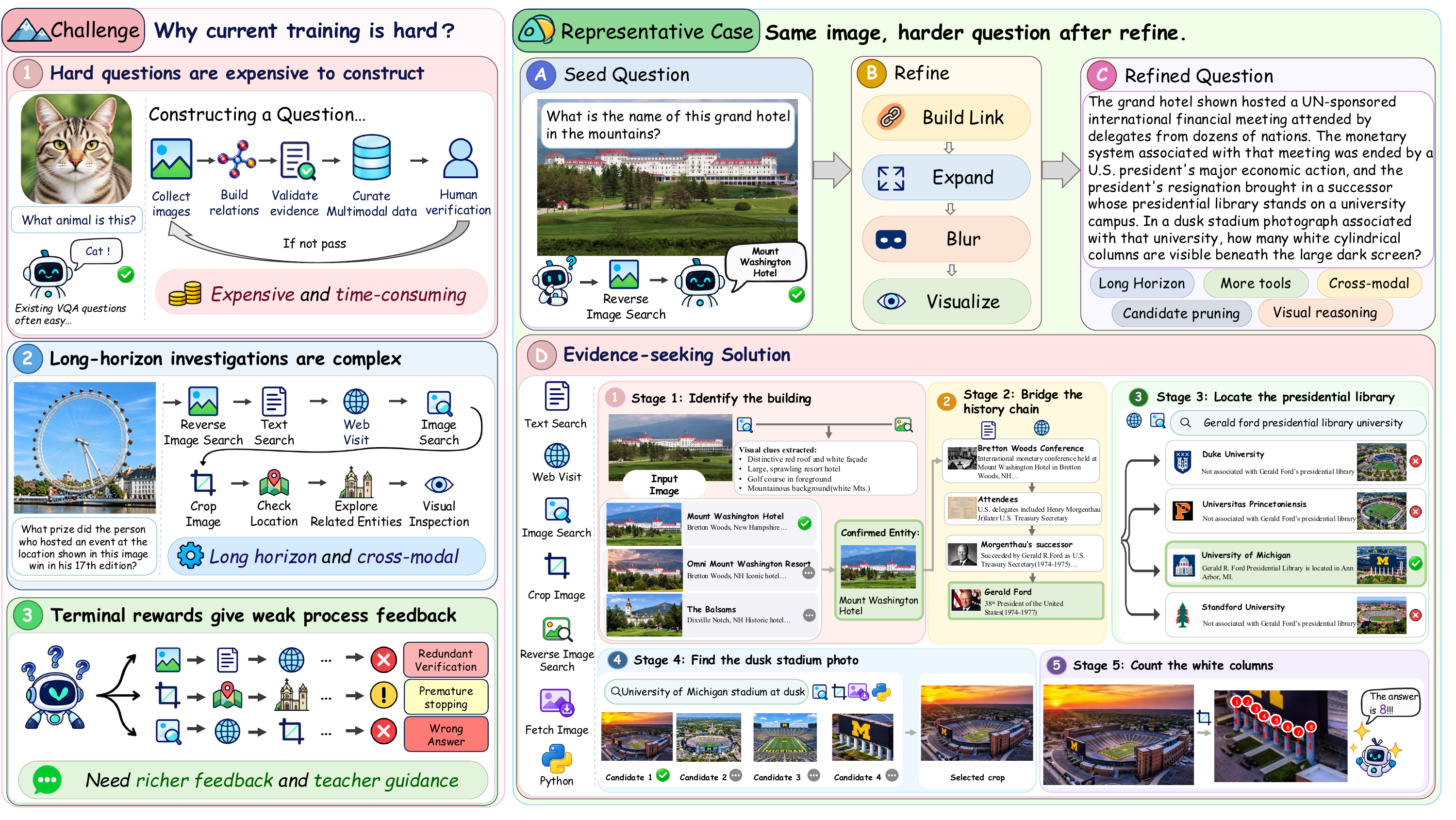}
  \caption{Motivation for MMSearch-OPD and an example of Refine for long-horizon multimodal search.}
  \label{fig:refine-comparison}
\end{figure*}

Multimodal deep search agents address knowledge-intensive visual questions
through interleaved reasoning and external retrieval
\citep{mmsearchr1,webwatcher}. Recent advances move toward
long-horizon investigation, connecting evidence across webpages
and using acquired visual observations to guide subsequent searches
\citep{visiondeepresearch,lmmsearcher,zhang2026deepvoyagervlincentivizingvisionintheloopsearch}.

Two challenges limit training for this setting. Existing multimodal
datasets contain many questions solvable through parametric knowledge or
short retrieval, while constructing harder questions through graph
composition and visual enrichment entails relation collection,
image curation, and evidence validation
\citep{chen2026opensearchvl,redsearcher,zhang2026deepvoyagervlincentivizingvisionintheloopsearch}.
Reusing existing image--question pairs offers a starting point for
constructing questions with greater search demands.
Meanwhile, useful long-horizon investigation may end in redundant
verification, premature stopping, or an incorrect answer. Terminal
rewards provide limited process feedback, and trajectory-level
advantages supply a shared scalar signal throughout a rollout
\citep{shao2024deepseekmath}. Identical returns also remove
group-relative learning contrast. Process rewards and teacher
guidance address these limitations
\citep{chen2026opensearchvl,li2026hypereyes}; we combine outcome-based
optimization with selective token-level teacher guidance.

To address these challenges, we propose \textbf{MMSearch-OPD},
combining controllable question refinement with
joint reinforcement learning (RL) and on-policy distillation (OPD).
Our \textbf{Refine} pipeline reuses simple multimodal questions and their
candidate reasoning chains.
\textbf{Expand} extends chain depth, \textbf{Blur} increases
disambiguation requirements, and \textbf{Visualize} introduces
retrievable visual evidence. These configurable operators control
the intended evidence requirements while reusing seed assets
(Figure~\ref{fig:refine-comparison}). After question
validation and supervised initialization, joint training combines
answer rewards and behavior penalties with token-level OPD on
imperfect student trajectories. The shaped reward gap adjusts
guidance strength, while an exponential schedule reduces the OPD
coefficient as training proceeds.

Our contributions are summarized as follows:
\begin{itemize}
  \item We introduce Refine, a pipeline that reuses existing
  multimodal questions and adjusts their evidence requirements
  through chain expansion, entity disambiguation, and visual
  evidence requirements.
  \item We develop joint RL and selective OPD training that combines
  answer rewards and behavior penalties with reward-gap-weighted
  token guidance and exponential annealing on imperfect
  long-horizon search trajectories.
  \item We organize evaluation across ten multimodal search
  benchmarks to assess agent performance and the roles of data
  refinement and joint training.
\end{itemize}

\section{Related Work}
\label{sec:related-work}

\paragraph{Long-Horizon Multimodal Search.}
MMSearch-R1 and WebWatcher train multimodal agents to interleave
reasoning with external retrieval \citep{mmsearchr1,webwatcher}.
Recent work extends search horizons and manages growing interaction
histories \citep{visiondeepresearch,lmmsearcher,pointsseeker}.
DeepVoyager-VL uses intermediate visual evidence to guide subsequent
search
\citep{zhang2026deepvoyagervlincentivizingvisionintheloopsearch}.
These developments call for training questions requiring sustained,
cross-modal evidence acquisition.

\paragraph{Multimodal Search Data Construction.}
Questions solvable through parametric knowledge or short retrieval
provide limited supervision for long-horizon search. Existing
pipelines increase difficulty through multi-hop composition and
entity obfuscation \citep{visiondeepresearch,chen2026opensearchvl},
graph-structured synthesis \citep{redsearcher,mei2026searchart}, and
visual evidence
integration
\citep{visualseeker,zhang2026deepvoyagervlincentivizingvisionintheloopsearch}.
These constructions require evidence collection and validation.
Our Refine pipeline reuses existing easy questions and candidate
reasoning chains, applying Expand, Blur, and Visualize to control
chain length, disambiguation, and visual requirements without
building a large graph from scratch.

\paragraph{Reinforcement Learning for Multimodal Search.}
GRPO-based training improves search policies through interaction
\citep{shao2024deepseekmath,visiondeepresearch}, but terminal scores
offer limited feedback on partially useful trajectories.
Process rewards, adaptive sampling, and finer credit assignment
address this limitation
\citep{chen2026opensearchvl,yan2026prommsearchagent,simplesearchvl,
jiao2026searcheyesfrontiermultimodaldeep,mei2025aisearchplanner,
zhang2025adamixt}.
HyperEyes combines efficiency rewards with teacher distillation
on failed rollouts \citep{li2026hypereyes}. Our joint optimization
combines answer rewards and behavior penalties with reward-gap-weighted
teacher guidance on imperfect student trajectories, using exponential
annealing to reduce reliance on the SFT teacher during training.

\begin{figure*}[!t]
  \centering
  \includegraphics[width=0.95\textwidth]{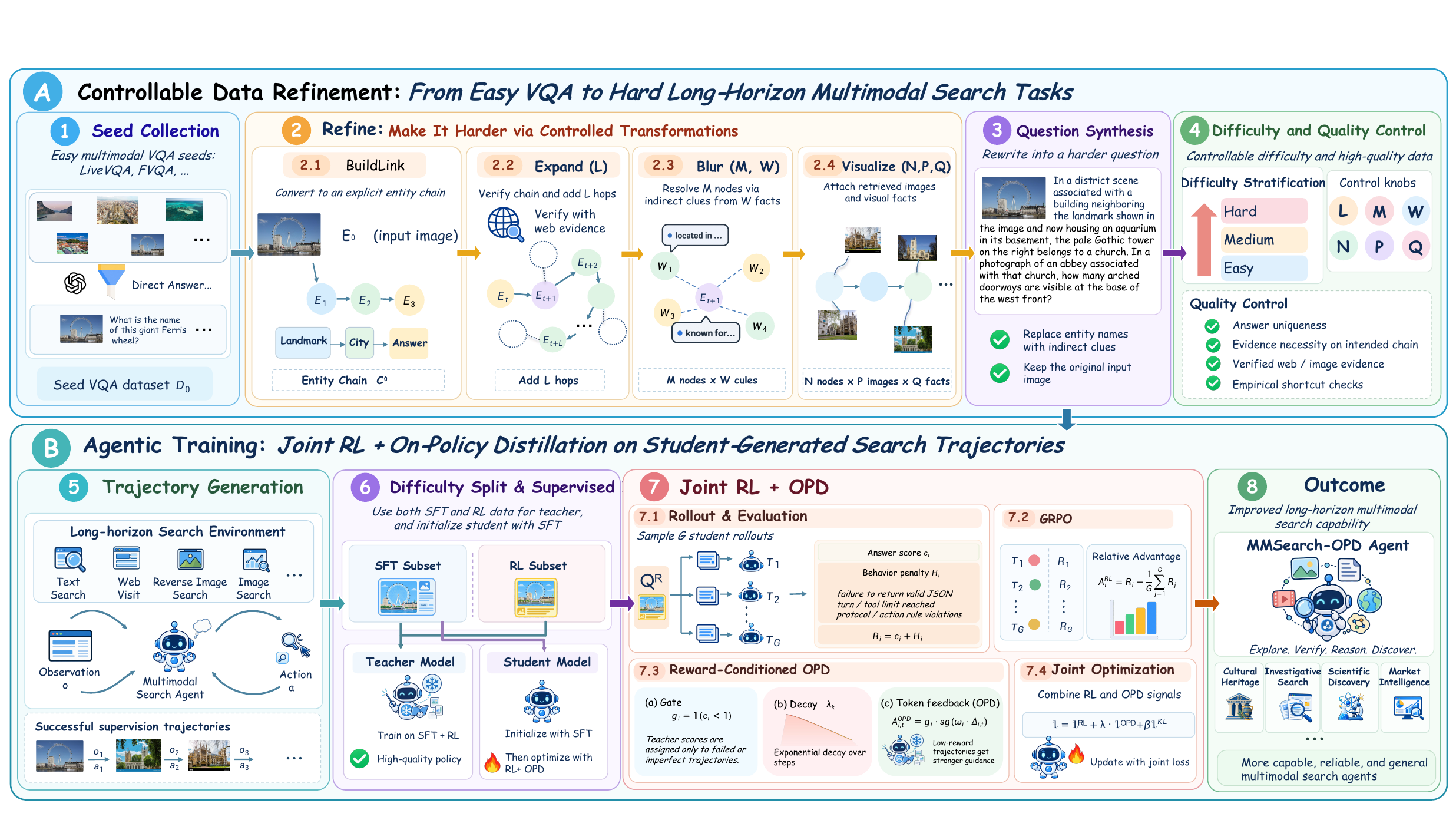}
  \caption{Overview of MMSearch-OPD: data refinement and joint RL--OPD training.}
  \label{fig:refine-pipeline}
\end{figure*}

\section{Method}
\label{sec:method}

MMSearch-OPD combines controllable question refinement with joint
RL--OPD training. Refine increases textual and visual evidence
requirements; after validation and supervised initialization,
outcome rewards and behavior penalties guide RL, while selective,
reward-gap-weighted token guidance supports learning on imperfect
student trajectories.

\subsection{Problem Formulation}
\label{sec:problem-formulation}

\paragraph{Long-horizon multimodal search.}
Given a question--image pair $x=(q,\mathcal{I})$, an agent
$\pi_\theta$ interleaves reasoning with web search and image
inspection until returning an answer. Later actions depend on the
accumulated multimodal history $h_t$, making evidence acquisition
sequential.
For example, identifying an entity in a retrieved image can determine
the next textual query; facts found on a webpage can instead guide
which image to inspect.

\paragraph{Reasoning chains and evidence necessity.}
We represent a task's designated reasoning route as an entity chain
$C^0=(E_0,\ldots,E_T)$: $E_0$ is grounded in the input image, each
clue $\ell_i$ links $E_{i-1}$ to $E_i$, and $E_T$ is the answer.
Each transition requires identifying the next entity from its
predecessor and linking clue.
Let $\mathcal{Z}$ contain the retrieved textual and visual facts
needed along this chain, and $\mathcal{Y}(x,\mathcal{Z})$ the answers
consistent with them. Our curation criterion requires a unique
answer and necessity of every designated evidence item:
\begin{equation}
\begin{gathered}
\mathcal{Y}(x,\mathcal{Z})=\{y\},\\
\mathcal{Y}(x,\mathcal{Z}\setminus\{z\})\neq\{y\},
\quad\forall z\in\mathcal{Z}.
\end{gathered}
\label{eq:evidence-necessity}
\end{equation}
Omitting an item therefore leaves the answer ambiguous or underivable
along this route.
This criterion applies to the intended chain; filtering in
Section~\ref{sec:quality-control} checks empirically for shortcuts
through parametric knowledge or alternative web evidence.

\subsection{Data Curation}
\label{sec:data-curation}

\paragraph{Seed collection.}
We take the full training splits of LiveVQA and FVQA and retain
questions that the data-generation model answers correctly without
tools. These seeds are easy for this model, and their successful
chain-of-thought provides candidate reasoning chains.

\subsubsection{Refine Pipeline}
\label{sec:refine-pipeline}
BuildLink extracts a seed's entity chain; an open-source model equipped
with web-search and image tools then applies Expand, Blur, and
Visualize in this order, followed by question synthesis
(Figure~\ref{fig:refine-pipeline}).
Six parameters $(L,M,W,N,P,Q)$ control the added evidence requirements.
Expand increases chain depth, Blur introduces entity disambiguation,
and Visualize adds retrieved visual evidence. The operators modify
the designated chain before question wording is synthesized, allowing
their evidence requirements to be controlled separately.

\paragraph{BuildLink.}
For a retained seed $x^0=(q^0,\mathcal{I})$ and its successful
chain-of-thought $r^0$, BuildLink extracts the candidate entity chain:
\begin{equation}
C^0=\textsc{BuildLink}(x^0,r^0)=(E_0,\ldots,E_T).
\label{eq:buildlink}
\end{equation}
Each transition records the linking clue $\ell_i$ from $E_{i-1}$
to $E_i$. This removes free-form reasoning artifacts and exposes
the entities and relations on which refinement operates. The root
$E_0$ is grounded in the input image, and $E_T$ is the original
answer; Expand verifies the extracted transitions against web evidence.

\paragraph{Expand.}
Expand verifies every original transition against web evidence,
discards invalid chains, and appends exactly $L$ evidence-supported
hops:
\begin{equation}
\begin{aligned}
C^{\mathrm{exp}}&=\textsc{Expand}_{L}(C^0)\\
&=(E_0,\ldots,E_T,E_{T+1},\ldots,E_{T+L}).
\end{aligned}
\label{eq:expand-op}
\end{equation}
Each appended entity has a web-grounded linking clue. The original
answer $E_T$ becomes an intermediate result, and $E_{T+L}$ the new
answer target.
Thus the solver must continue searching beyond the seed's answer
while following the verified prefix of the chain.

\paragraph{Blur.}
Blur selects $M$ internal nodes, indexed by $\Omega_B$, and attaches
$W$ distinct, source-grounded facts
$\mathcal{B}_i=\{\beta_{i,1},\ldots,\beta_{i,W}\}$ to each. These
facts jointly identify $E_i$ and are attached to the selected node:
\begin{equation}
\begin{aligned}
E_i^{\mathrm{b}}&=(E_i,\mathcal{B}_i),\quad i\in\Omega_B,\\
C^{\mathrm{blur}}&=\textsc{Blur}_{M,W}(C^{\mathrm{exp}}).
\end{aligned}
\label{eq:blur-op}
\end{equation}
Entities and chain topology remain unchanged. Question synthesis
later uses $\mathcal{B}_i$ to replace the explicit entity name with
indirect, searchable clues.
Resolving these descriptions adds entity disambiguation within an
existing hop.

\paragraph{Visualize.}
Visualize selects $N$ nodes, indexed by $\Omega_V$, from the terminal
node backward. For each, it retains $P_i\leq P$ retrieved images
$\mathcal{J}_{i,p}$ and extracts $Q$ task-relevant visual facts
$\nu_{i,p,q}$ per image. Images are retained only when the target
can be localized and the visual facts verified. The resulting evidence is
\begin{equation}
\mathcal{V}_i=\bigl\{(\mathcal{J}_{i,p},\{\nu_{i,p,q}\}_{q=1}^{Q})\bigr\}_{p=1}^{P_i}.
\label{eq:visual-evidence}
\end{equation}
It is attached as $E_i^{\mathrm{v}}=(E_i,\mathcal{B}_i,\mathcal{V}_i)$,
with $\mathcal{B}_i=\emptyset$ if the node was not blurred. The
resulting refined chain is
\begin{equation}
\widetilde C=\textsc{Visualize}_{N,P,Q}(C^{\mathrm{blur}}).
\label{eq:visualize-op}
\end{equation}
Terminal-node visualization supports a pixel-grounded answer;
earlier-node visualization adds visual evidence at intermediate hops.
Blur and Visualize can affect original or appended nodes, including
the same node, and add evidence requirements within hops rather than
necessarily adding tool turns.

\paragraph{Question synthesis.}
We first draft a question with explicit entity names tracing
$\widetilde C$ from the input image to $E_{T+L}$. We then replace names at nodes
in $\Omega_B\cup\Omega_V$ with descriptions based on
$\mathcal{B}_i$ and $\mathcal{V}_i$, requiring the solver to resolve
textual clues and retrieve and inspect the associated images.
This two-stage synthesis preserves the chain, answer, and images:
\begin{equation}
x^R=(q^R,\mathcal{I})=\textsc{Synthesis}(\widetilde C).
\label{eq:synthesis}
\end{equation}
For a visualized node, the description requires locating the associated
image before the entity can be identified, connecting visual evidence
to the subsequent search.

\subsubsection{Quantifying Search Difficulty}
\label{sec:theory}
Let $K(z\mid h)$ count the minimum new evidence items needed to
uniquely identify $z$ given context $h$, and $D_{\mathrm{init}}$ sum
these requirements along the original chain. Refinement gives
\begin{equation}
\begin{aligned}
D_{\mathrm{Refined}}
={}&D_{\mathrm{init}}
+\underbrace{\textstyle\sum_{i=T+1}^{T+L}K(E_i\mid \mathcal{H}_i)}_{\text{Expand}}\\
&+\underbrace{\textstyle\sum_{i\in\Omega_B}\sum_{w=1}^{W}
K(\beta_{i,w}\mid \mathcal{H}_{i,w}^{\mathrm{blur}})}_{\text{Blur}}\\
&+\underbrace{\textstyle\sum_{i\in\Omega_V}\sum_{p=1}^{P_i}
K^{\mathrm{vis}}_{i,p}}_{\text{Visualize}},
\end{aligned}
\label{eq:refine-icrc}
\end{equation}
where $\mathcal{H}_i=(E_{i-1},\ell_i)$ is the pre-search hop context,
$\mathcal{H}_{i,w}^{\mathrm{blur}}$ contains node $i$'s clue and previously
resolved Blur facts, $P_i\leq P$ is the number of retained images,
and $K^{\mathrm{vis}}_{i,p}$ accounts for image
retrieval, verification, and reading its $Q$ visual facts.
Each evidence item is an atomic, verifiable textual or visual fact.
Selecting the same node for Blur and Visualize contributes to both
only when its textual and visual requirements are distinct.
This accounting assumes that added evidence remains necessary given
earlier evidence and that shared facts are not counted twice.
Positive added terms increase the designated chain's evidence burden;
the measure does not specify a global minimum search route, tool-turn
count, or model success rate.

\subsection{Quality Control and Trajectory Generation}
\label{sec:quality-control}

\paragraph{Question validation.}
Each synthesized question is walked through from the input image,
with minimal wording fixes to ensure solvability and answer uniqueness.
We check chain consistency, evidence provenance, image--text alignment,
and actual use of the designated Blur and Visualize evidence. These
checks approximate Equation~\ref{eq:evidence-necessity} and help ensure
that the added requirements in Equation~\ref{eq:refine-icrc} are
non-trivial.

\paragraph{Difficulty filtering and trajectory selection.}
We discard questions answered correctly by a direct-answer probe
without tools. An expert model then performs $k$ tool-augmented
rollouts per question using the training search-and-vision interface.
The probe removes residual parametric shortcuts that refinement alone
cannot rule out, while tool-augmented rollouts test whether the
questions are learnable through external evidence acquisition.
The expert's empirical Pass@$k$ measures agent difficulty, complementing the
chain-based evidence burden $D_{\mathrm{Refined}}$.
We retain successful trajectories as SFT demonstrations, sampling
across Pass@$k$ levels to balance difficulty and prevent easy
questions from dominating supervision. These estimates also guide
the SFT/RL split described in Section~\ref{sec:training}.

\subsection{Agentic Training}
\label{sec:training}

\paragraph{Supervised initialization.}
Using the difficulty estimates in Section~\ref{sec:quality-control},
we partition the curated training questions into an SFT subset and
a harder RL subset. A larger model is fine-tuned on successful
trajectories from both subsets and frozen as teacher
$\pi_{\mathrm{T}}$. The student $\pi_\theta$ is initialized by
SFT on the first subset and jointly optimized with RL and OPD on
the second.

\paragraph{Joint RL and OPD for long-horizon search.}
Useful evidence collection can end in redundant verification,
premature stopping, or an unsupported answer. Final-answer scores
provide limited feedback on these intermediate decisions, motivating
guidance along the search process.
We combine outcome-based policy optimization with token-level
teacher guidance on the student's own trajectories, supplementing
the shared scalar feedback from trajectory-wide advantages.
For each question, we sample $G$ student rollouts and assign
correctness scores $c_i\in[0,1]$. The shaped reward and group-centered
GRPO advantage
\citep{shao2024deepseekmath} are
\begin{equation}
\begin{aligned}
R_i&=c_i+H_i,\\
A_i^{\mathrm{RL}}&=R_i-\frac{1}{G}\sum_{j=1}^{G}R_j,
\end{aligned}
\label{eq:process-advantage}
\end{equation}
where $H_i\le0$ penalizes inefficient tool usage and repetitive
search behavior.

\paragraph{Reward-conditioned guidance.}
The teacher scores each sampled token $u_{i,t}$ conditioned on
the student's multimodal history $h_{i,t}$
\citep{agarwal2024onpolicy,li2026hypereyes}. For trajectories with $c_i<1$ and
valid teacher scores, set $g_i=1$; otherwise $g_i=0$.
Within this subset, the shaped reward determines guidance strength
through $\omega_i=[1-R_i]_+$. The token feedback is
\begin{equation}
\begin{aligned}
\Delta_{i,t}
&=\operatorname{clip}\!\left(
\log\frac{\pi_{\mathrm{T}}(u_{i,t}\mid h_{i,t})}
{\pi_\theta(u_{i,t}\mid h_{i,t})},
-c_{\mathrm{opd}},c_{\mathrm{opd}}\right),\\
A_{i,t}^{\mathrm{OPD}}
&=g_i\operatorname{sg}(\omega_i\Delta_{i,t}).
\end{aligned}
\label{eq:opd-token-advantage}
\end{equation}
Here $\operatorname{sg}$ denotes stop-gradient. Guidance can vary
in sign across tokens, remain nonzero when group advantages vanish,
and apply to failed rollouts with positive RL advantage. Correctness
controls eligibility, while the shaped reward gap controls strength.
Being better than its peers does not imply that a rollout reaches the
target reward. Fully correct trajectories receive no OPD, even when
behavior penalties lower their shaped reward; RL still encourages
more efficient search.
We minimize the combined objective
\begin{equation}
\mathcal L(\theta)=\mathcal L_{\mathrm{RL}}
+\lambda_s\mathcal L_{\mathrm{OPD}}+\beta\mathcal L_{\mathrm{KL}},
\label{eq:joint-training}
\end{equation}
where $\mathcal L_{\mathrm{KL}}$ regularizes against a fixed reference
policy with $\beta=0.01$. We exponentially decay $\lambda_s$ from
0.05 by a factor of 100 every ten RL steps, reducing imitation
pressure as outcome-based optimization progresses. Early guidance
is intended to supplement weak group-relative signals from frequent
failures; its decay leaves room for the student to move beyond the
SFT teacher's preferences.

\begin{table*}[!t]
\centering
\begingroup
\definecolor{MMTableSectionGray}{HTML}{ECEFF3}
\definecolor{MMTableScaleGray}{HTML}{F5F6F8}
\definecolor{MMTableAverageGray}{HTML}{F5F6F8}
\definecolor{MMTableAccent}{HTML}{244D73}
\definecolor{MMTableSFTTint}{HTML}{F0F6FB}
\definecolor{MMTableFinalTint}{HTML}{E2EDF8}
\definecolor{MMTableAverageAccent}{HTML}{CCDFF1}
\newcommand{\mmTableHeader}[1]{\textbf{#1}}
\newcommand{\mmTableCite}[1]{{\fontsize{6.8}{7.5}\selectfont\citep{#1}}}

\fontsize{8.5}{10.1}\selectfont
\setlength{\tabcolsep}{3.5pt}
\renewcommand{\arraystretch}{1.02}
\begin{adjustbox}{max width=\textwidth}
\begin{tabular}{l*{7}{c}>{\columncolor{MMTableAverageGray}}c}
\toprule[0.8pt]
\mmTableHeader{Model} & \mmTableHeader{MMSearch} & \mmTableHeader{LiveVQA} & \mmTableHeader{BC-VL} & \mmTableHeader{MM-BC} & \mmTableHeader{MMSearch+} & \mmTableHeader{BC-V$^3$} & \mmTableHeader{VisBrowse} & \mmTableHeader{Avg.} \\
\midrule[0.35pt]
\rowcolor{MMTableSectionGray}\multicolumn{9}{c}{\rule{0pt}{10pt}\bfseries Direct Answer} \\
GPT-5.5~\mmTableCite{openai2026gpt55} & \textbf{68.7} & \textbf{73.0} & \textbf{47.9} & \textbf{17.5} & \underline{20.3} & \textbf{23.0} & \textbf{36.1} & \textbf{40.9} \\
Gemini-3.1-Pro~\mmTableCite{google2026gemini31} & \underline{64.2} & 65.0 & 41.4 & 11.5 & \textbf{26.4} & \underline{19.3} & 23.7 & \underline{35.9} \\
Claude-Opus-4.6~\mmTableCite{anthropic2026opus46} & 59.8 & 53.1 & 43.5 & 13.2 & 13.2 & 15.0 & 27.2 & 32.1 \\
Seed2.1-Pro~\mmTableCite{bytedance2026seed21} & 53.7 & \underline{68.7} & \underline{45.6} & \underline{14.3} & 17.7 & 17.0 & \underline{27.8} & 35.0 \\
Qwen3.5-27B~\mmTableCite{qwen2026qwen35} & 29.3 & 56.7 & 32.3 & 5.0 & 4.8 & 9.3 & 14.8 & 21.7 \\
Qwen3.5-9B~\mmTableCite{qwen2026qwen35} & 19.0 & 52.0 & 25.1 & 3.25 & 2.3 & 7.3 & 11.2 & 17.2 \\
\midrule[0.35pt]
\rowcolor{MMTableSectionGray}\multicolumn{9}{c}{\rule{0pt}{10pt}\bfseries Agentic Workflow} \\
GPT-5.5~\mmTableCite{openai2026gpt55} & \textbf{82.7} & \textbf{90.3} & \textbf{68.2} & \underline{51.9} & 48.6 & \underline{55.0} & \underline{66.3} & \underline{66.1} \\
Gemini-3.1-Pro~\mmTableCite{google2026gemini31} & \textbf{82.7} & 87.3 & 65.9 & 44.4 & 51.5 & 50.0 & 62.1 & 63.4 \\
Claude-Opus-4.6~\mmTableCite{anthropic2026opus46} & 81.7 & 88.0 & 63.2 & 49.4 & \underline{52.1} & \underline{55.0} & 62.7 & 64.6 \\
Seed2.1-Pro~\mmTableCite{bytedance2026seed21} & \underline{82.2} & \underline{89.7} & \underline{67.9} & \textbf{60.8} & \textbf{62.2} & \textbf{61.7} & \textbf{72.8} & \textbf{71.0} \\
Qwen3.5-27B~\mmTableCite{qwen2026qwen35} & \textbf{82.7} & 87.0 & 59.2 & 38.5 & 38.9 & 39.7 & 50.9 & 56.7 \\
Qwen3.5-9B~\mmTableCite{qwen2026qwen35} & 75.7 & 79.0 & 49.9 & 25.3 & 29.9 & 23.7 & 30.8 & 44.9 \\
\midrule[0.5pt]
\rowcolor{MMTableSectionGray}\multicolumn{9}{c}{\rule{0pt}{10pt}\bfseries Multimodal Deep Search Agents} \\
\rowcolor{MMTableScaleGray}\multicolumn{9}{l}{\rule{0pt}{10pt}\bfseries\itshape Small-scale (7B--9B)} \\
MMSearch-R1-7B~\mmTableCite{mmsearchr1} & 53.8 & 48.4 & -- & -- & -- & 4.0 & -- & -- \\
WebWatcher-7B~\mmTableCite{webwatcher} & 49.1 & 51.2 & 21.2 & -- & -- & 4.7 & -- & -- \\
DeepEyesV2-7B~\mmTableCite{deepeyesv2} & 63.7 & -- & -- & -- & -- & -- & -- & -- \\
Vision-DeepResearch-8B~\mmTableCite{visiondeepresearch} & 69.6 & 76.7 & 42.6 & -- & 20.4 & 11.7 & -- & -- \\
OpenSearch-VL-8B~\mmTableCite{chen2026opensearchvl} & 64.5 & 59.6 & 37.6 & -- & -- & -- & -- & -- \\
SimpleSearch-VL-8B~\mmTableCite{simplesearchvl} & 77.1 & 75.2 & 52.1 & -- & 32.5 & -- & -- & -- \\
Visual-Seeker-8B~\mmTableCite{visualseeker} & 72.2 & -- & 47.6 & 16.1 & 27.3 & -- & 34.7 & -- \\
DeepVoyager-VL-8B~\mmTableCite{zhang2026deepvoyagervlincentivizingvisionintheloopsearch} & 72.7 & 82.7 & 58.4 & 24.0 & 37.1 & 32.3 & 47.3 & 50.6 \\
\addlinespace[2pt]
\rowcolor{MMTableSFTTint}\textcolor{MMTableAccent}{\textbf{MMSearch-OPD-9B (SFT)}} & \underline{85.3} & \underline{87.3} & \underline{65.7} & \underline{38.8} & \underline{45.7} & \underline{33.0} & \underline{49.7} & \cellcolor{MMTableSFTTint}\underline{57.9} \\
\rowcolor{MMTableFinalTint}\textcolor{MMTableAccent}{\textbf{MMSearch-OPD-9B (RL)}} & \textbf{86.3} & \textbf{89.0} & \textbf{67.7} & \textbf{42.0} & \textbf{47.0} & \textbf{37.3} & \textbf{52.1} & \cellcolor{MMTableAverageAccent}\textbf{60.2} \\
\midrule[0.5pt]
\rowcolor{MMTableScaleGray}\multicolumn{9}{l}{\rule{0pt}{10pt}\bfseries\itshape Large-scale (27B--32B)} \\
WebWatcher-32B~\mmTableCite{webwatcher} & 55.3 & 58.7 & 27.0 & -- & -- & 8.7 & -- & -- \\
Vision-DeepResearch-30B-A3B~\mmTableCite{visiondeepresearch} & 69.6 & 77.6 & 53.7 & -- & 28.5 & -- & -- & -- \\
REDSearcher-MM-30B-A3B\textsuperscript{$\dagger$}~\mmTableCite{redsearcher} & 72.9 & 79.3 & 57.2 & 23.5 & 26.6 & -- & -- & -- \\
OpenSearch-VL-30B-A3B~\mmTableCite{chen2026opensearchvl} & 68.7 & 67.4 & 41.1 & -- & -- & -- & -- & -- \\
HyperEyes-30B-A3B~\mmTableCite{li2026hypereyes} & \textbf{86.9} & 81.6 & 57.9 & -- & 31.5 & -- & -- & -- \\
SimpleSearch-VL-30B-A3B~\mmTableCite{simplesearchvl} & 83.6 & 81.1 & 55.9 & -- & 34.4 & -- & -- & -- \\
DeepVoyager-VL-30B-A3B~\mmTableCite{zhang2026deepvoyagervlincentivizingvisionintheloopsearch} & 74.0 & \underline{82.7} & \underline{64.2} & \underline{30.5} & \underline{40.6} & \underline{35.0} & \underline{53.8} & \underline{54.4} \\
\addlinespace[2pt]
\rowcolor{MMTableSFTTint}\textcolor{MMTableAccent}{\textbf{MMSearch-OPD-27B (SFT)}} & \underline{86.3} & \textbf{90.0} & \textbf{69.4} & \textbf{52.3} & \textbf{44.4} & \textbf{45.0} & \textbf{55.0} & \cellcolor{MMTableSFTTint}\textbf{63.2} \\
\bottomrule[0.8pt]
\end{tabular}
\end{adjustbox}
\endgroup
\caption{Performance (\%) on seven multimodal search benchmarks. \textbf{Bold}/\underline{underline}: best/second-best within each setting and scale. Avg.: seven-benchmark mean; --: unreported. \textsuperscript{$\dagger$}: Thinking initialization.}
\label{tab:main-result}
\end{table*}

\section{Experiments}
\label{sec:experiments}

\subsection{Experimental Setup}
\label{sec:experimental-setup}

\paragraph{Models and Training Data.}
We use Seed2.1-Pro~\citep{bytedance2026seed21} to generate SFT
trajectories. Qwen3.5-27B is fine-tuned on all curated trajectories
and frozen as the OPD teacher. Qwen3.5-9B receives SFT on a
subset, followed by joint RL--OPD
training on harder questions. We report its SFT and RL checkpoints.

\paragraph{Benchmarks and Evaluation.}
We evaluate on MMSearch~\citep{mmsearch}, LiveVQA~\citep{livevqa},
BrowseComp-VL~\citep{webwatcher}, MM-BrowseComp~\citep{mmbrowsecomp},
MMSearch-Plus~\citep{mmsearchplus}, BrowseComp-V\textsuperscript{3}~\citep{browsecompv3},
and VisBrowse-Bench~\citep{visbrowse}.
We use Qwen3.6-35B-A3B to judge final-answer correctness with the
Vision-DeepResearch judge prompt~\citep{visiondeepresearch}.
Each prediction is judged three times, and the scores are averaged.
Avg. is the unweighted benchmark mean.

\paragraph{Baselines.}
We compare tool-free \emph{Direct Answer}, tool-enabled
\emph{Agentic Workflow}, and trained \emph{Multimodal Deep Search
Agents}, grouping the latter by model scale.

\paragraph{Implementation Details.}
We use ms-swift for SFT and veRL for joint RL--OPD training.
The agent interleaves reasoning with textual and visual retrieval,
with a maximum of 50 interaction turns per evaluation question.

\subsection{Main Results}
\label{sec:main-results}

\begin{figure*}[t]
\centering
\captionsetup{skip=4pt}
\begin{minipage}[t]{0.49\textwidth}
  \centering
  \includegraphics[width=\linewidth]{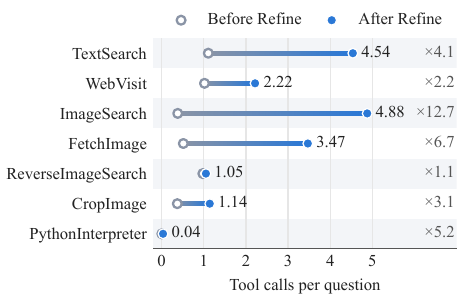}
  \par\vspace{1pt}
  {\small (a) Tool calls per question}
\end{minipage}
\hfill
\begin{minipage}[t]{0.49\textwidth}
  \centering
  \includegraphics[width=\linewidth]{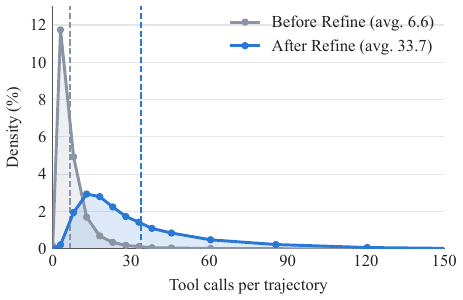}
  \par\vspace{1pt}
  {\small (b) Tool-call distribution}
\end{minipage}
\caption{Tool-use statistics before and after Refine. (a) counts tool
invocations. (b) counts each search query separately and plots the share
of trajectories per tool call; dashed lines mark the averages.}
\label{fig:refine-statistics}
\end{figure*}

\paragraph{Overall Performance.}
Table~\ref{tab:main-result} shows that MMSearch-OPD-9B reaches
60.2 average accuracy, improving over the corresponding Qwen3.5-9B
agentic workflow by 15.3 points. It achieves the highest reported
accuracy among the compared 7B--9B search agents on all seven
benchmarks, exceeding DeepVoyager-VL-8B by 9.6 points on average.
The 27B SFT teacher reaches 63.2, a 6.5-point improvement over its
agentic-workflow baseline. The trained 9B student also exceeds the
untuned 27B workflow by 3.5 points on average.

\paragraph{Long-Horizon Multimodal Search.}
MMSearch-OPD improves performance across the evaluated search
settings, gaining 12.8 points on average over the same 9B agentic
workflow on MMSearch, LiveVQA, and BrowseComp-VL.
In particular, it gains 17.2 points on average on MM-BrowseComp,
MMSearch-Plus, BrowseComp-V\textsuperscript{3}, and VisBrowse-Bench,
which place substantial demands on long-horizon investigation and
cross-modal evidence integration. These results demonstrate
substantial improvements in search settings requiring sustained
evidence acquisition and joint reasoning over text and images.

\paragraph{Joint Training Beyond SFT.}
The 9B SFT checkpoint already improves average accuracy from
44.9 to 57.9. Subsequent joint RL--OPD training, labeled RL in
Table~\ref{tab:main-result}, further raises it to 60.2, with
improvements on every benchmark. The 2.3-point average gain
supports joint online optimization after SFT.

\subsection{Ablation Study}
\label{sec:ablation}

\subsubsection{Data Refinement Ablation}
\label{sec:data-refinement-ablation}

We pair refined questions with their original seeds and use
Seed2.1-Pro~\citep{bytedance2026seed21} to generate both sets of
trajectories. The two Qwen3.5-9B models use identical initialization,
SFT configurations, and training budgets; only question refinement
differs. Table~\ref{tab:data-refinement-ablation} reports SFT results
on SimpleVQA~\citep{simplevqa}, FVQA~\citep{wang2017fvqa},
VDR~\citep{vdrbench}, and BrowseComp-V\textsuperscript{3}~\citep{browsecompv3}.

\begin{table}[t]
\centering
\begingroup
\definecolor{MMRefineTint}{HTML}{E2EDF8}
\definecolor{MMRefineAccent}{HTML}{244D73}
\fontsize{8.5}{10.1}\selectfont
\setlength{\tabcolsep}{3pt}
\renewcommand{\arraystretch}{1.08}
\begin{adjustbox}{max width=\linewidth}
\begin{tabular}{lcccc}
\toprule[0.8pt]
\textbf{Training Data} & \textbf{SimpleVQA} & \textbf{FVQA} & \textbf{VDR} & \textbf{BC-V$^3$} \\
\midrule[0.35pt]
Before Refine & 72.7 & 74.0 & 29.0 & 18.0 \\
\rowcolor{MMRefineTint}
\textcolor{MMRefineAccent}{\textbf{After Refine}} & \textbf{79.0} & \textbf{76.0} & \textbf{35.8} & \textbf{33.0} \\
$\Delta$ & +6.3 & +2.0 & +6.8 & +15.0 \\
\bottomrule[0.8pt]
\end{tabular}
\end{adjustbox}
\endgroup
\caption{Effect of Refine on Qwen3.5-9B after SFT. $\Delta$ denotes accuracy gains in percentage points.}
\label{tab:data-refinement-ablation}
\end{table}

Refine improves all four benchmarks, with the largest gain on
BrowseComp-V\textsuperscript{3} (+15.0 points), which requires
long-horizon browsing with interleaved textual and visual evidence.
VDR, SimpleVQA, and FVQA improve by 6.8, 6.3, and 2.0 points,
respectively, supporting the training value of refined questions.

Figure~\ref{fig:refine-statistics} shows average tool calls per
trajectory increasing from 6.61 to 33.71, counting each search query
separately. Per question, ImageSearch and FetchImage invocations
rise from 0.38 to 4.88 and from 0.51 to 3.47, respectively;
the distribution shifts toward longer searches. Together with
the SFT gains, these statistics indicate more sustained search
and greater visual evidence acquisition after Refine.

\begin{figure*}[t]
\centering
\captionsetup{skip=4pt}
\begin{minipage}[t]{0.49\textwidth}
  \centering
  \includegraphics[width=0.85\linewidth]{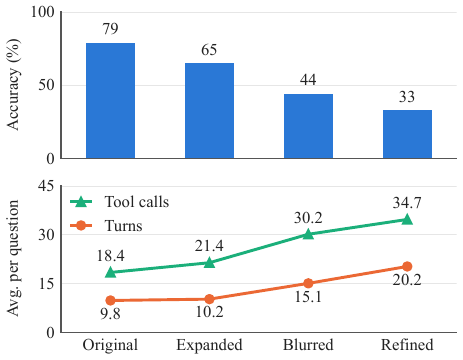}
  \par\vspace{1pt}
  {\small (a) Accuracy and search length}
\end{minipage}
\hfill
\begin{minipage}[t]{0.49\textwidth}
  \centering
  \includegraphics[width=0.85\linewidth]{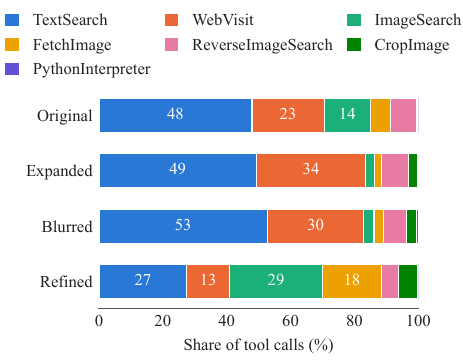}
  \par\vspace{1pt}
  {\small (b) Tool-use composition}
\end{minipage}
\caption{Refine pipeline ablation on 100 paired questions.}
\label{fig:refine-pipeline-ablation}

\par\medskip
\begin{minipage}[t]{0.49\textwidth}
  \centering
  \includegraphics[width=0.85\linewidth]{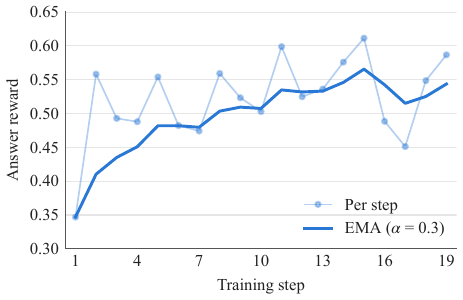}
  \par\vspace{1pt}
  {\small (a) Answer reward}
\end{minipage}
\hfill
\begin{minipage}[t]{0.49\textwidth}
  \centering
  \includegraphics[width=0.85\linewidth]{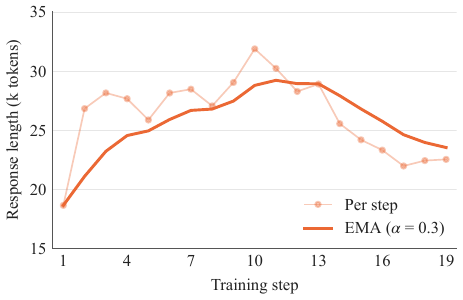}
  \par\vspace{1pt}
  {\small (b) Response length}
\end{minipage}
\caption{RL training dynamics.}
\label{fig:rl-training-dynamics}
\end{figure*}

\subsubsection{Refine Pipeline Ablation}
\label{sec:stage-wise-refine}
\label{sec:refine-pipeline-ablation}

We follow the same 100 seed questions through Expand, Blur, and
Visualize, synthesizing a question from each intermediate structure.
Seed2.1-Pro solves all four versions under identical rollout
settings. Figure~\ref{fig:refine-pipeline-ablation} reports accuracy,
search length, and tool-use composition. Each search request within
a batched action counts as one tool call.

Each step strengthens the questions' search requirements:
accuracy decreases from 79\% to 65\%, 44\%, and 33\%.
Expand lowers accuracy with little change in interaction length;
Blur increases average turns from 10.2 to 15.1, consistent with
additional entity disambiguation. Visualize further raises turns
to 20.2 and tool calls to 34.7 per question, while ImageSearch and
FetchImage's combined share of tool actions increases from 6.3\%
to 47.5\%. These results support the complementary roles of chain
expansion, disambiguation, and visual evidence acquisition in
constructing harder long-horizon multimodal search tasks.

\subsection{RL Training Analysis}
\label{sec:rl-training-analysis}

\paragraph{Reward Dynamics.}
Figure~\ref{fig:rl-training-dynamics}(a) shows mean answer reward
before behavior penalties rising from 0.347 to 0.587 over 19 steps.
Despite batch fluctuations, this trend indicates improved answer
quality during joint RL--OPD training.

\paragraph{Search Behavior.}
Search trajectories initially lengthen and then contract
(Figure~\ref{fig:rl-training-dynamics}(b)). Comparing steps 6--13
with steps 14--19, average tool calls decrease from 26.0 to 21.3
(18.3\%), and generated response length falls from 29.0k to 23.4k
tokens (19.5\%), while mean answer reward remains comparable
(0.525 versus 0.544). These observations are consistent with
increasingly economical search: later rollouts use fewer
interactions and generated tokens without a corresponding decline
in average answer quality. These batch-level trends describe the
joint training run rather than isolate the effect of individual
components.

\section{Conclusion}
\label{sec:conclusion}

We introduced MMSearch-OPD, combining controllable data refinement
with joint RL--OPD training for long-horizon multimodal deep search.
Refine reuses existing multimodal question--image pairs and applies targeted
transformations to control search depth, entity disambiguation,
and visual evidence requirements. Joint training complements
answer rewards and behavior penalties with reward-conditioned
token-level guidance on imperfect student trajectories, while
progressively reducing teacher influence. Across seven primary
benchmarks, our 9B agent improves average accuracy by 15.3
percentage points over its agentic-workflow baseline, with joint
training adding 2.3 points beyond SFT. Data ablations further
support the training value of refined questions. Together, these
results show the value of upgrading existing multimodal data to
require sustained evidence acquisition. They also support joint
outcome optimization and selective teacher guidance for learning
long-horizon search and reasoning over text and images.

\section*{Limitations}

Refine increases evidence requirements along designated reasoning
chains, but filtering cannot rule out all shortcuts from parametric
knowledge or alternative web evidence. Training depends on teacher
and answer-judge quality, which may introduce noisy guidance.
Open-web search also relies on changing content and tool
availability, limiting reproducibility.

\bibliography{custom}

@article{chen2026opensearchvl,
  title={OpenSearch-VL: An Open Recipe for Frontier Multimodal Search Agents},
  author={Chen, Shuang and Feng, Kaituo and Chen, Hangting and Huang, Wenxuan and Dai, Dasen and Shou, Quanxin and Lin, Yunlong and Yue, Xiangyu and Gao, Shenghua and Pang, Tianyu},
  journal={arXiv preprint arXiv:2605.05185},
  year={2026}
}

@article{li2026hypereyes,
  title={HyperEyes: Dual-Grained Efficiency-Aware Reinforcement Learning for Parallel Multimodal Search Agents},
  author={Li, Guankai and Chen, Jiabin and Xu, Yi and Zhang, Xichen and Lu, Yuan},
  journal={arXiv preprint arXiv:2605.07177},
  year={2026}
}

@article{shao2024deepseekmath,
    author  = {Zhihong Shao and Peiyi Wang and Qihao Zhu and Runxin Xu and Junxiao Song and Xiao Bi and Haowei Zhang and Mingchuan Zhang and Y. K. Li and Y. Wu and Daya Guo},
    title   = {{DeepSeekMath}: Pushing the Limits of Mathematical Reasoning in Open Language Models},
    journal = {arXiv preprint arXiv:2402.03300},
    year    = {2024},
    url     = {https://arxiv.org/abs/2402.03300}
}

@inproceedings{agarwal2024onpolicy,
    author    = {Rishabh Agarwal and Nino Vieillard and Yongchao Zhou and Piotr Stanczyk and Sabela Ramos Garea and Matthieu Geist and Olivier Bachem},
    title     = {On-Policy Distillation of Language Models: Learning from Self-Generated Mistakes},
    booktitle = {The Twelfth International Conference on Learning Representations},
    year      = {2024},
    url       = {https://arxiv.org/abs/2306.13649}
}

@misc{mmsearchr1,
      title={MMSearch-R1: Incentivizing LMMs to Search},
      author={Jinming Wu and Zihao Deng and Wei Li and Yiding Liu and Bo You and Bo Li and Zejun Ma and Ziwei Liu},
      year={2025},
      eprint={2506.20670},
      archivePrefix={arXiv},
      primaryClass={cs.CV},
      url={https://arxiv.org/abs/2506.20670},
}

@misc{webwatcher,
      title={WebWatcher: Breaking New Frontier of Vision-Language Deep Research Agent},
      author={Xinyu Geng and Peng Xia and Zhen Zhang and Xinyu Wang and Qiuchen Wang and Ruixue Ding and Chenxi Wang and Jialong Wu and Yida Zhao and Kuan Li and Yong Jiang and Pengjun Xie and Fei Huang and Jingren Zhou},
      year={2025},
      eprint={2508.05748},
      archivePrefix={arXiv},
      primaryClass={cs.IR},
      url={https://arxiv.org/abs/2508.05748},
}

@misc{visiondeepresearch,
      title={Vision-DeepResearch: Incentivizing DeepResearch Capability in Multimodal Large Language Models},
      author={Wenxuan Huang and Yu Zeng and Qiuchen Wang and Zhen Fang and Shaosheng Cao and Zheng Chu and Qingyu Yin and Shuang Chen and Zhenfei Yin and Lin Chen and Zehui Chen and Xu Tang and Yao Hu and Shaohui Lin and Philip Torr and Feng Zhao and Wanli Ouyang},
      year={2026},
      eprint={2601.22060},
      archivePrefix={arXiv},
      primaryClass={cs.CV},
      url={https://arxiv.org/abs/2601.22060},
}

@misc{lmmsearcher,
      title={Towards Long-horizon Agentic Multimodal Search},
      author={Yifan Du and Zikang Liu and Jinbiao Peng and Jie Wu and Junyi Li and Jinyang Li and Wayne Xin Zhao and Ji-Rong Wen},
      year={2026},
      eprint={2604.12890},
      archivePrefix={arXiv},
      primaryClass={cs.CV},
      url={https://arxiv.org/abs/2604.12890},
}

@misc{pointsseeker,
      title={POINTS-Seeker: Towards Training a Multimodal Agentic Search Model from Scratch},
      author={Yikun Liu and Yuan Liu and Le Tian and Xiao Zhou and Jiangchao Yao and Yanfeng Wang and Weidi Xie},
      year={2026},
      eprint={2604.14029},
      archivePrefix={arXiv},
      primaryClass={cs.CV},
      url={https://arxiv.org/abs/2604.14029},
}

@misc{redsearcher,
      title={REDSearcher: A Scalable and Cost-Efficient Framework for Long-Horizon Search Agents},
      author={Zheng Chu and Xiao Wang and Jack Hong and Huiming Fan and Yuqi Huang and Yue Yang and Guohai Xu and Chenxiao Zhao and Cheng Xiang and Shengchao Hu and Dongdong Kuang and Ming Liu and Bing Qin and Xing Yu},
      year={2026},
      eprint={2602.14234},
      archivePrefix={arXiv},
      primaryClass={cs.AI},
      url={https://arxiv.org/abs/2602.14234},
}

@misc{visualseeker,
      title={Visual-Seeker: Towards Visual-Native Multimodal Agentic Search via Active Visual Reasoning},
      author={Zhengbo Zhang and Changtao Miao and Jinbo Su and Zhaowen Zhou and Chunxia Zhang and Xukai Wang and Ruiqi Liu and Kaiyuan Zheng and Jiansheng Cai and Bo Zhang and Zhe Li and Shiming Xiang and Ying Yan},
      year={2026},
      eprint={2606.15231},
      archivePrefix={arXiv},
      primaryClass={cs.AI},
      url={https://arxiv.org/abs/2606.15231},
}

@misc{simplesearchvl,
      title={SimpleSearch-VL: A Simple Recipe for Multimodal Agentic Deep Search},
      author={Ming Dai and Zhihong Lu and Jinjie Gu and Jiedong Zhuang and Yefeng Liu and Wankou Yang and Jian Wang and Chunhua Shen},
      year={2026},
      eprint={2606.31504},
      archivePrefix={arXiv},
      primaryClass={cs.CV},
      url={https://arxiv.org/abs/2606.31504},
}

@article{yan2026prommsearchagent,
  title={ProMMSearchAgent: A Generalizable Multimodal Search Agent Trained with Process-Oriented Rewards},
  author={Yan, Wentao and Wang, Shengqin and Zhou, Huichi and Chen, Yihang and Shao, Kun and Xie, Yuan and Zhang, Zhizhong},
  journal={arXiv preprint arXiv:2604.20486},
  year={2026}
}

@misc{zhang2026deepvoyagervlincentivizingvisionintheloopsearch,
      title={{DeepVoyager-VL}: Incentivizing Vision-in-the-Loop Search for Long-Horizon Multimodal Agents},
      author={Huanyao Zhang and Jiepeng Zhou and Runhao Zhao and Yanzhe Shan and Jiaoyang Chen and Bowen Zhou and Bo Li and Fang Wang and Jialong Wu and Zhengwei Tao and Lang Mei and Xiaohan Yu and Liyan Liu and Chong Chen and Wentao Zhang},
      year={2026},
      eprint={2608.01827},
      archivePrefix={arXiv},
      primaryClass={cs.CV},
      url={https://arxiv.org/abs/2608.01827},
}

@misc{jiao2026searcheyesfrontiermultimodaldeep,
      title={SearchEyes: Towards Frontier Multimodal Deep Search Intelligence via Search World Simulation},
      author={Zhengbo Jiao and Yiming Cheng and Yilei Jiang and Kaituo Feng and Rui Huang and Tianyi Jiang and Juanxi Tian and Jiapeng li and Qunzhong Wang and Tailai Chen and Qianshan Wei and Chuan Xiao and Shanyu Rong and Yangfu Li and Yanhan Zhou and Yunpu Ma and Yifan Zhang and Xiangyu Yue},
      year={2026},
      eprint={2607.05943},
      archivePrefix={arXiv},
      primaryClass={cs.AI},
      url={https://arxiv.org/abs/2607.05943},
}

@inproceedings{mmsearch,
      title={{MMSearch}: Benchmarking the Potential of Large Models as Multi-modal Search Engines},
      author={Dongzhi Jiang and Renrui Zhang and Ziyu Guo and Yanmin Wu and Jiayi Lei and Pengshuo Qiu and Pan Lu and Zehui Chen and Chaoyou Fu and Guanglu Song and Peng Gao and Yu Liu and Chunyuan Li and Hongsheng Li},
      booktitle={The Thirteenth International Conference on Learning Representations},
      year={2025},
}

@misc{livevqa,
      title={Seeking and Updating with Live Visual Knowledge},
      author={Mingyang Fu and Yuyang Peng and Dongping Chen and Zetong Zhou and Benlin Liu and Yao Wan and Zhou Zhao and Philip S. Yu and Ranjay Krishna},
      year={2025},
      eprint={2504.05288},
      archivePrefix={arXiv},
      primaryClass={cs.CV},
      url={https://arxiv.org/abs/2504.05288},
}

@misc{mmbrowsecomp,
      title={MM-BrowseComp: A Comprehensive Benchmark for Multimodal Browsing Agents},
      author={Shilong Li and Xingyuan Bu and Wenjie Wang and Jiaheng Liu and Jun Dong and Haoyang He and Hao Lu and Haozhe Zhang and Chenchen Jing and Zhen Li and Chuanhao Li and Jiayi Tian and Chenchen Zhang and Tianhao Peng and Yancheng He and Jihao Gu and Yuanxing Zhang and Jian Yang and Ge Zhang and Wenhao Huang and Wangchunshu Zhou and Zhaoxiang Zhang and Ruizhe Ding and Shilei Wen},
      year={2025},
      eprint={2508.13186},
      archivePrefix={arXiv},
      primaryClass={cs.CL},
      url={https://arxiv.org/abs/2508.13186},
}

@misc{mmsearchplus,
      title={MMSearch-Plus: Benchmarking Provenance-Aware Search for Multimodal Browsing Agents},
      author={Xijia Tao and Yihua Teng and Xinxing Su and Xinyu Fu and Jihao Wu and Chaofan Tao and Ziru Liu and Haoli Bai and Rui Liu and Lingpeng Kong},
      year={2026},
      eprint={2508.21475},
      archivePrefix={arXiv},
      primaryClass={cs.AI},
      url={https://arxiv.org/abs/2508.21475},
}

@misc{browsecompv3,
      title={BrowseComp-$V^3$: A Visual, Vertical, and Verifiable Benchmark for Multimodal Browsing Agents},
      author={Huanyao Zhang and Jiepeng Zhou and Bo Li and Bowen Zhou and Yanzhe Shan and Haishan Lu and Zhiyong Cao and Jiaoyang Chen and Yuqian Han and Zinan Sheng and Zhengwei Tao and Hao Liang and Jialong Wu and Yang Shi and Yuanpeng He and Jiaye Lin and Qintong Zhang and Guochen Yan and Runhao Zhao and Zhengpin Li and Xiaohan Yu and Lang Mei and Chong Chen and Wentao Zhang and Bin Cui},
      year={2026},
      eprint={2602.12876},
      archivePrefix={arXiv},
      primaryClass={cs.AI},
      url={https://arxiv.org/abs/2602.12876},
}

@misc{visbrowse,
      title={VisBrowse-Bench: Benchmarking Visual-Native Search for Multimodal Browsing Agents},
      author={Zhengbo Zhang and Jinbo Su and Zhaowen Zhou and Changtao Miao and Yuhan Hong and Qimeng Wu and Yumeng Liu and Feier Wu and Yihe Tian and Yuhao Liang and Zitong Shan and Wanke Xia and Yi-Fan Zhang and Bo Zhang and Zhe Li and Shiming Xiang and Ying Yan},
      year={2026},
      eprint={2603.16289},
      archivePrefix={arXiv},
      primaryClass={cs.CV},
      url={https://arxiv.org/abs/2603.16289},
}

@misc{deepeyesv2,
      title={DeepEyesV2: Toward Agentic Multimodal Model},
      author={Jack Hong and Chenxiao Zhao and ChengLin Zhu and Weiheng Lu and Guohai Xu and Xing Yu},
      year={2026},
      eprint={2511.05271},
      archivePrefix={arXiv},
      primaryClass={cs.CV},
      url={https://arxiv.org/abs/2511.05271},
}

@misc{openai2026gpt55,
  title={{GPT-5.5}},
  author={{OpenAI}},
  year={2026},
  url={https://developers.openai.com/api/docs/models/gpt-5.5}
}

@misc{google2026gemini31,
  title={{Gemini 3.1 Pro}: A Smarter Model for Your Most Complex Tasks},
  author={{Google DeepMind}},
  year={2026},
  url={https://blog.google/innovation-and-ai/models-and-research/gemini-models/gemini-3-1-pro/}
}

@misc{anthropic2026opus46,
  title={Introducing {Claude Opus 4.6}},
  author={{Anthropic}},
  year={2026},
  url={https://www.anthropic.com/news/claude-opus-4-6}
}

@misc{bytedance2026seed21,
  title={{Seed2.1} Officially Released: Advancing {AI} Productivity},
  author={{ByteDance Seed}},
  year={2026},
  url={https://seed.bytedance.com/en/blog/seed2-1-officially-released-advancing-ai-productivity}
}

@misc{qwen2026qwen35,
  title={{Qwen3.5}: Towards Native Multimodal Agents},
  author={{Qwen Team}},
  year={2026},
  url={https://qwen.ai/blog?id=qwen3.5}
}

@inproceedings{simplevqa,
      title={{SimpleVQA}: Multimodal Factuality Evaluation for Multimodal Large Language Models},
      author={Xianfu Cheng and Wei Zhang and Shiwei Zhang and Jian Yang and Xiangyuan Guan and Xianjie Wu and Xiang Li and Ge Zhang and Jiaheng Liu and Yuying Mai and Yutao Zeng and Zhoufutu Wen and Ke Jin and Baorui Wang and Weixiao Zhou and Yunhong Lu and Tongliang Li and Wenhao Huang and Zhoujun Li},
      booktitle={Proceedings of the IEEE/CVF International Conference on Computer Vision},
      pages={4637--4646},
      year={2025},
}

@misc{vdrbench,
      title={Vision-DeepResearch Benchmark: Rethinking Visual and Textual Search for Multimodal Large Language Models},
      author={Yu Zeng and Wenxuan Huang and Zhen Fang and Shuang Chen and Yufan Shen and Yishuo Cai and Xiaoman Wang and Zhenfei Yin and Lin Chen and Zehui Chen and Shiting Huang and Yiming Zhao and Xu Tang and Yao Hu and Philip Torr and Wanli Ouyang and Shaosheng Cao},
      year={2026},
      eprint={2602.02185},
      archivePrefix={arXiv},
      primaryClass={cs.CV},
      url={https://arxiv.org/abs/2602.02185},
}

@article{wang2017fvqa,
  title={Fvqa: Fact-based visual question answering},
  author={Wang, Peng and Wu, Qi and Shen, Chunhua and Dick, Anthony and Van Den Hengel, Anton},
  journal={IEEE transactions on pattern analysis and machine intelligence},
  volume={40},
  number={10},
  pages={2413--2427},
  year={2017},
  publisher={IEEE}
}

@misc{mei2026searchart,
  title={{SearchArt}: Training Long-Horizon Search Agent with Scalable Synthetic and Verified Task},
  author={Lang Mei and Xiaohan Yu and Chong Chen and Liyan Liu and Xiangnan Chen and Jinchao Ma and Chao Feng and Li Huang and Siyu Mo and Sichen Kang and Yunkun Xu and Zhihan Yang and Zhujun Xue and Jingren Zhang and Qing He and Yingdi Huang and Hao Jiang and Ziao Ma and Zewei Pan and Minhao Sun and Zhuo Tao and Jinzhao Xiao and Gangtao Xin and Huanyao Zhang and Wenjian Zhang and Jiangshan Zhang and Guojie Zhu and Fangzhou Zou and Jiaxin Mao and Wentao Zhang},
  year={2026},
  eprint={2607.24850},
  archivePrefix={arXiv},
  primaryClass={cs.IR},
  url={https://arxiv.org/abs/2607.24850}
}

@misc{mei2025aisearchplanner,
  title={{AI-SearchPlanner}: Modular Agentic Search via Pareto-Optimal Multi-Objective Reinforcement Learning},
  author={Lang Mei and Zhihan Yang and Xiaohan Yu and Huanyao Zhang and Chong Chen},
  year={2025},
  eprint={2508.20368},
  archivePrefix={arXiv},
  primaryClass={cs.AI},
  url={https://arxiv.org/abs/2508.20368}
}

@misc{zhang2025adamixt,
  title={{AdaMixT}: Adaptive Weighted Mixture of Multi-Scale Expert Transformers for Time Series Forecasting},
  author={Huanyao Zhang and Jiaye Lin and Wentao Zhang and Haitao Yuan and Guoliang Li},
  year={2025},
  eprint={2509.18107},
  archivePrefix={arXiv},
  primaryClass={cs.LG},
  url={https://arxiv.org/abs/2509.18107}
}

\end{document}